\documentclass{styles/svproc}
\usepackage{url}

\usepackage{todonotes}
\usepackage[normalem]{ulem}
\usepackage{caption}
\usepackage{subcaption}

\usepackage{array}
\usepackage{booktabs}
\newcolumntype{M}[1]{>{\centering\arraybackslash}m{#1}}
\begin{document}
\mainmatter              
\title{Defining Adaptive Proxemic Zones for Activity-aware Navigation}
\titlerunning{Proxemic Zones for Activity-aware}  

\author{Jonatan Gin\'es Clavero\inst{1} \and Francisco Mart\'in Rico\inst{2}
Francisco J. Rodr\'iguez-Lera\inst{3} \and Jos\'e Miguel Guerrero Hern\'andez\inst{2} \and Vicente Matell\'an Olivera\inst{4}}

\authorrunning{Jonatan Gin\'es et al.} 

\institute{Escuela Internacional de Doctorado - Rey Juan Carlos University (Spain).\\
\email{j.gines@alumnos.urjc.es}
\and
Intelligent Robotics Lab, Rey Juan Carlos University (Spain). \\
\and
Escuela de Ingenier\'ias Industrial e Inform\'atica, Universidad de Le\'on (Spain).\\
\and
Supercomputaci\'on Castilla y Le\'on, SCAYLE, Le\'on (Spain)}

\maketitle              

\begin{abstract}
Many of the tasks that a service robot can perform at home involve navigation skills. In a real world scenario, the navigation system should consider individuals beyond just objects, theses days it is necessary to offer particular and dynamic representation in the scenario in order to enhance the HRI experience. In this paper, we use the proxemic theory to do this representation. The proxemic zones are not static. The culture or the context influences them and, if we have this influence into account, we can increase humans' comfort. Moreover, there are collaborative tasks in which these zones take different shapes to allow the task's best performance. This research develops a layer, the \textit{social layer}, to represent and distribute the proxemics zones' information in a standard way, through a cost map and using it to perform a social navigate task. We have evaluated these components in a simulated scenario, performing different collaborative and human-robot interaction tasks and reducing the personal area invasion in a 32\%. The material developed during this research can be found in a public repository\footnote[1]{https://github.com/IntelligentRoboticsLabs/social\_navigation2\_WAF}, as well as instructions to facilitate the reproducibility of the results.
\end{abstract}

\section{Introduction}
\label{sec:intro}
A human sharing his home with a robot is getting closer every day and is starting to stop being a science-fiction movie thing. So far, domestic robots have a particular purpose, like the vacuum cleaner, but it is expected that in the coming years, these types of robots will have a general-purpose and will be able to solve everyday tasks, as well as naturally interact with humans. It requires robots to treat humans in a special way, as they will share space and tasks with them. Humans are letting be another obstacle that robots have to avoid, for example, during navigation.
More and more research is being done on \textit{social navigation}. This type of navigation takes into account humans, their social conventions, or their activity, improving their comfort \cite{KRUSE20131726}. One of the most used models in social navigation is the proxemics theory \cite{hall1910hidden}. It defines the space around a person as different zones with different radius: intimate, personal, social, and public. The intimate zone is the zone closest to the person (\textless 0.4m) and the personal zone (0.4 - 1.2m) 
In this work, we will focus on the intimate and personal areas. The intimate zone is an area that the robot must always respect, so navigation is forbidden in this zone. On the other hand, the personal zone is an area where the person interacts with known people or collaborates with others to perform a task. It can be a restricted navigation zone for the robots \cite{Vega2019}\cite{8206628}\cite{5326271} or, as described in this article, an adaptive zone. The personal zone can be considered adaptive because, depending on the context, it will be a restricted zone or a cooperation zone where the robot enters to carry out a task with the person.
In this way, we make the robot more natural, social, and adaptable. The system adapts to human-robot interaction, HRI, situations whereby robot and human being close or situations such as the current pandemic, caused by COVID-19, in which the safety distance of 2 meters must be respected. 

As already mentioned, the proxemics theory does not describe areas with a fixed radius but defines areas that could change according to the context, culture or age, among others. In \cite{vega2018aflexible}, the authors propose an approach in which the proxemic zones are dynamic and change depending on the spatial context and human intention. Another works have developed methods to follow the social convention of keep on the right when walking in a corridor using this theory as a base \cite{6696579}\cite{4107827}. However, our proposal is general to any scenario.

Another approach to develop a human-aware navigation is the use of a virtual forces model, the Social Force Model (SFM) \cite{Helbing_1995}. This approach consider the attractive virtual forces, created by the points of interest or the people who want to interact, and repulsive ones, created by the rest of the people or the obstacles.
Recent researches use a modern version of the SFM, the Extended Social Force Model \cite{Repiso2019}\cite{Galvan2020}. In these proposals, they have developed a planner and a controller based on this force model. Our proposal is independent of the planning and control algorithm used so it may be adapted to new and better navigation algorithms.

Another recent approach \cite{Koay2017} proposes the use of predefined positions around the person to interact with them. The positions are evaluated based on the user's preferences and choose the best position as the navigation goal. If that position is not reachable, the robot will go to the next best position. As this approximation does not represent the proxemic zones on the map, the robot can invade the intimate or personal zone while moving from one point to another, which would reduce the person's comfort. Our research tries to guarantee the comfort of the humans that interact or collaborate with the robot by establishing the intimate zone as a forbidden navigation zone.


Previous authors’ work proposed an initial social layer for the ROS navigation \cite{Gines2019}. It takes information about people's moods to adapt their proxemics zones, trying to do not disturb people with a bad mood. The paper at hand adapts this layer to the ROS2 navigation stack and extends the previous work to adapt the proxemics zones according to the context or the collaborative activity conducted by humans and robots. 
\subsection{Contribution}

The main contributions compared to previous works are divided into two, scientific and technical:
\begin{itemize}
    \item Scientific: A proxemic framework to represent, with adaptive proxemics shapes, human activities, location, culture, or specific situations.
    \item Scientific: A novel proxemic shape with the addition of what we call the \textit{cooperation zone}. It allows a fluid and natural cooperation between humans and robots in navigation and interaction tasks.
    \item Technical: The development of open-source ROS2 navigation layers, \textit{people filter}, and \textit{social layer}, brings the scientific community a standard and public framework to represent dynamic proxemics zones in a map and allow them to create complex behaviors using this work as a base.

\end{itemize}

\subsection{Paper Organization}
The work presented here presents and discusses both contributions. At first, section \ref{sec:Framework} defines the framework and how the proxemics zones are built. Section \ref{sec:Social_layer} defines the integration with ROS2 and how it works.
Section \ref{sec:Experiments} describes the performance of the system using the metrics established by the scientific community and presents an analysis of the experiment results and their implications. Finally, conclusions are presented in Section~\ref{sec:Conclusions}.
\section{Framework description}
\label{sec:Framework}

\begin{figure}[!ht]
\centering
\includegraphics[width=3.3in]{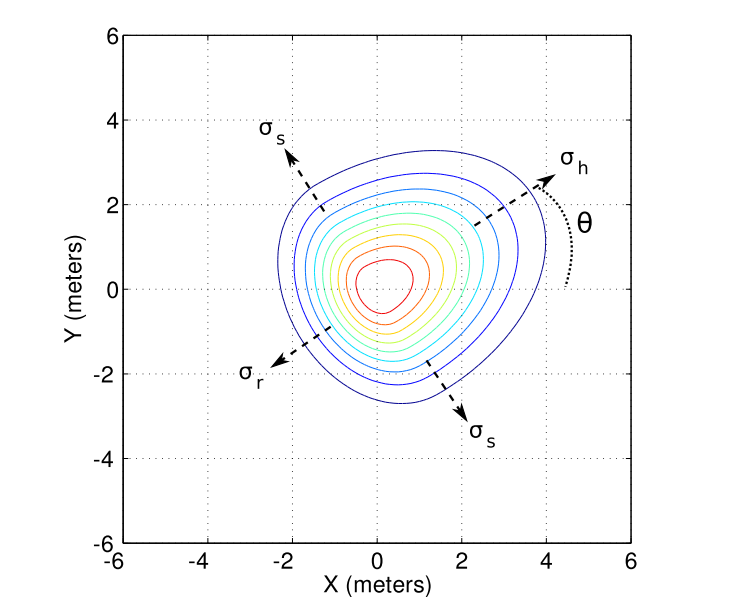}
\caption{Asymmetric Gaussian function centered at (0, 0), rotated by \(\Theta\) = \(\pi\)/6, and having variances \(\sigma_{h}\) = 2.0, \(\sigma_{s}\) = 4/3, and \(\sigma_{r}\) = 1.0. Figure A.1 from \cite{Kirby2010}.}
\label{fig:asymmetriGaussianKirby}
\end{figure}

This work proposes a framework for representing the space surrounding a person, the proxemic areas, on a cost map. This representation is fundamental to differentiate humans from the rest of the obstacles, thus enriching the robot's knowledge of its environment. Unlike past research that used proxemic zones based on Gaussian functions of concentric circles \cite{Gines2019}, in this article, the authors have used Asymmetric Gaussian proxemic zones \cite{Kirby2010}.
These proxemic zones provide us a high adaptation capacity to the context in which the robot is located, varying their size and shape, unlike the used in previous research that only modified their size. The Asymmetric Gaussian are defined by four variables: head (\(\sigma_{h}\)), side (\(\sigma_{s}\)), rear (\(\sigma_{r}\)) and an orientation (\(\Theta\)). Figure \ref{fig:asymmetriGaussianKirby} shows a graphic explanation of these parameters.

The high adaptability offered by this type of Gaussian allows us to associate different shapes and sizes of the proxemic zones with different activities of a human's daily life. Thus, associating some values for these variables with human activities, we can build a social map in which people are represented differently based on their activity. For example, people cooking, a person is moving at a certain speed or a person standing in the scenario, figure \ref{fig:activity_prox}.

Also, we propose new proxemics shapes oriented to improve in the perform of collaborative tasks between the human and the robot, taking as reference the work of Mead et al. \cite{Mead2015}. They show that humans adapt their proxemic zones to interact with a robot. In that way, we have designed proxemic zones that contain a \textit{cooperation zone}, figure \ref{fig:cooperation_zone_prox}. The robot will occupy the cooperation zone during the collaborative task to keep close to the person. These zones are located within the personal zone but always respecting the intimate zone.

\subsection{Asymmetric Gaussian function as human activity representation.}
The proposal for represent people and their activities uses the model described in \cite{Kirby2010}. In this model people generate areas where navigation is forbidden or penalised, using an asymmetric Gaussian function. Let P\(_{n}\) = \{p\(_{1}\), p\(_{2}\)...p\(_{n}\)\} be the set of n persons detected in the scenario and p\(_{i}\) = (x, y, \(\theta\)) is the pose of the person i. 

\begin{eqnarray*}
  g_{pi}(x,y) &=& e^{-(A(x-x_{i})^2+2B(x-x_{i})(y-y_{i})+C(y-y_{i})^2)}
\end{eqnarray*}

With A, B, C:
\begin{eqnarray*}
  A &=& \frac{cos(\theta)^2}{2\sigma^2} + \frac{sin(\theta)^2}{2\sigma_{s}^2}
\end{eqnarray*}
\begin{eqnarray*}
  B &=& \frac{sin(2\theta)}{4\sigma^2} - \frac{sin(2\theta)}{4\sigma_{s}^2}
\end{eqnarray*}
\begin{eqnarray*}
  C &=& \frac{sin(\theta)^2}{2\sigma^2} + \frac{cos(\theta)^2}{2\sigma_{s}^2}
\end{eqnarray*}
 where \(\sigma s\), as already mentioned, is the variance on the left and right.
\newcommand{\cookingFigure}{\includegraphics[width=2.3in]{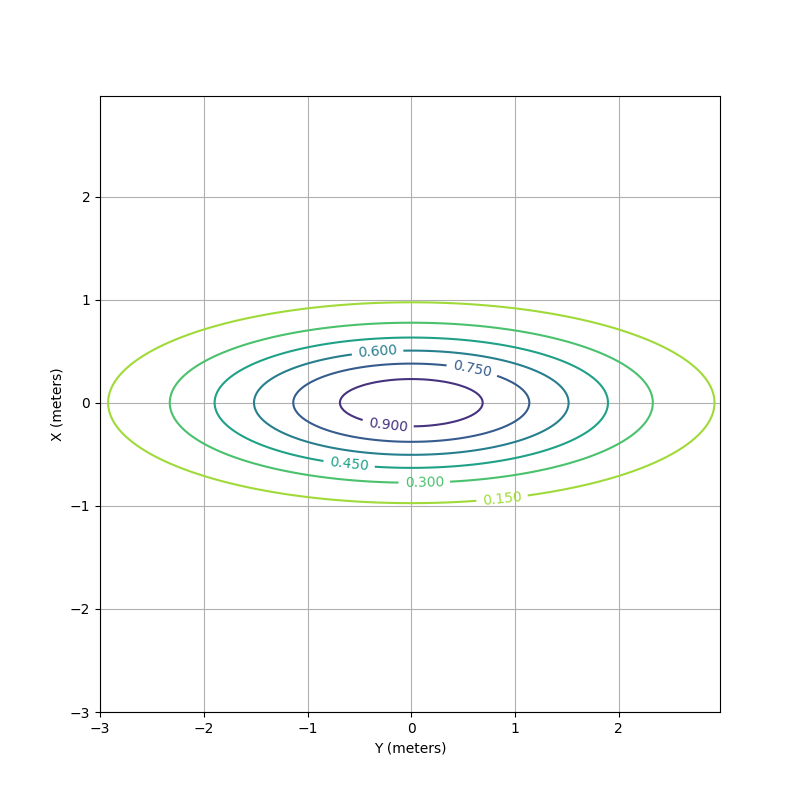}}
\newcommand{\runningFigure}{\includegraphics[width=2.3in]{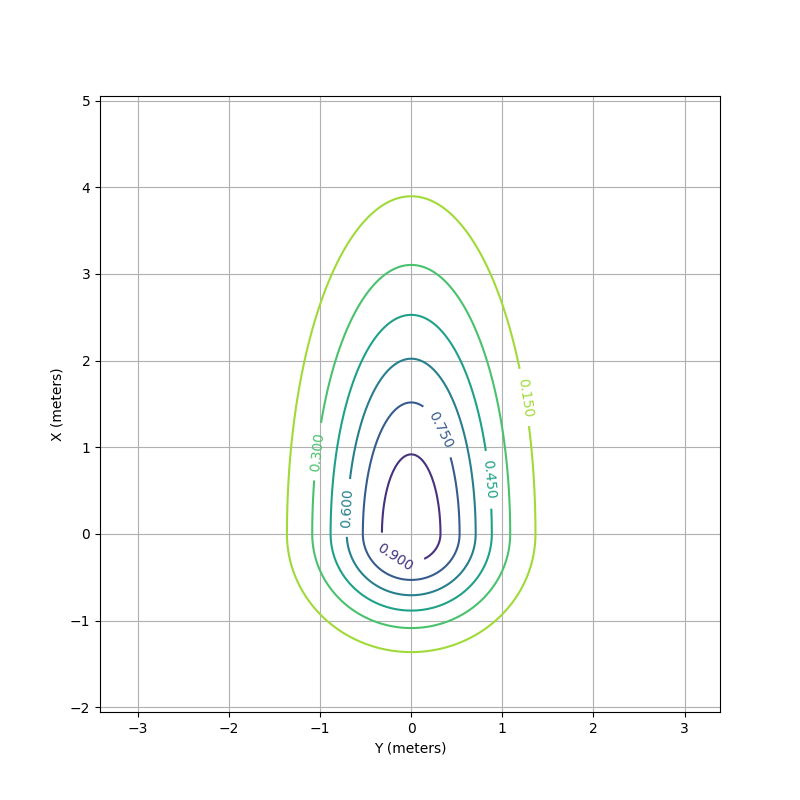}}
\newcommand{\standingFigure}{\includegraphics[width=2.3in]{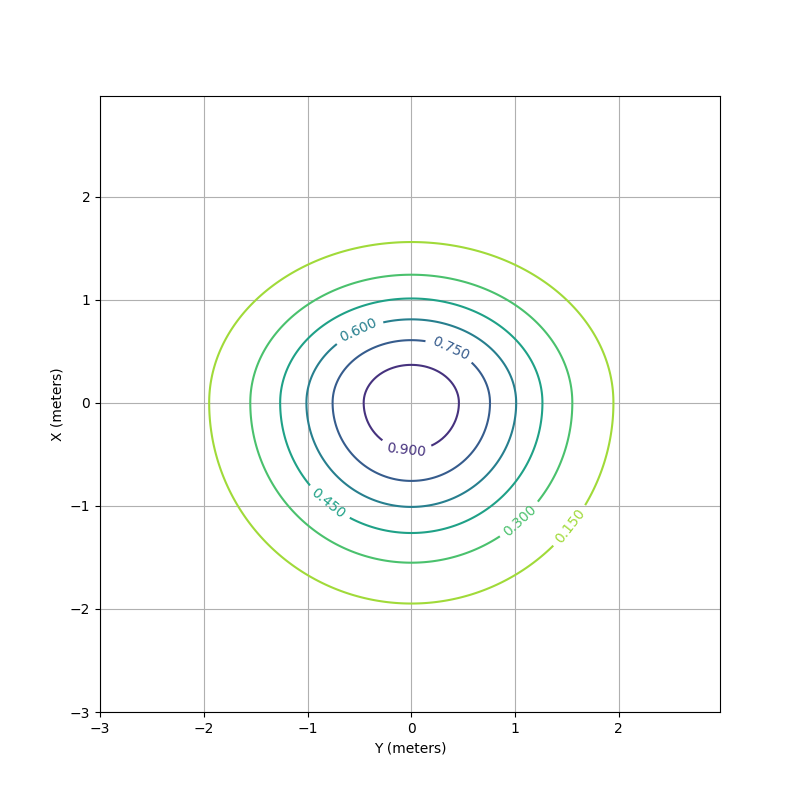}}
\newcommand{\bathroomFigure}{\includegraphics[width=2.3in]{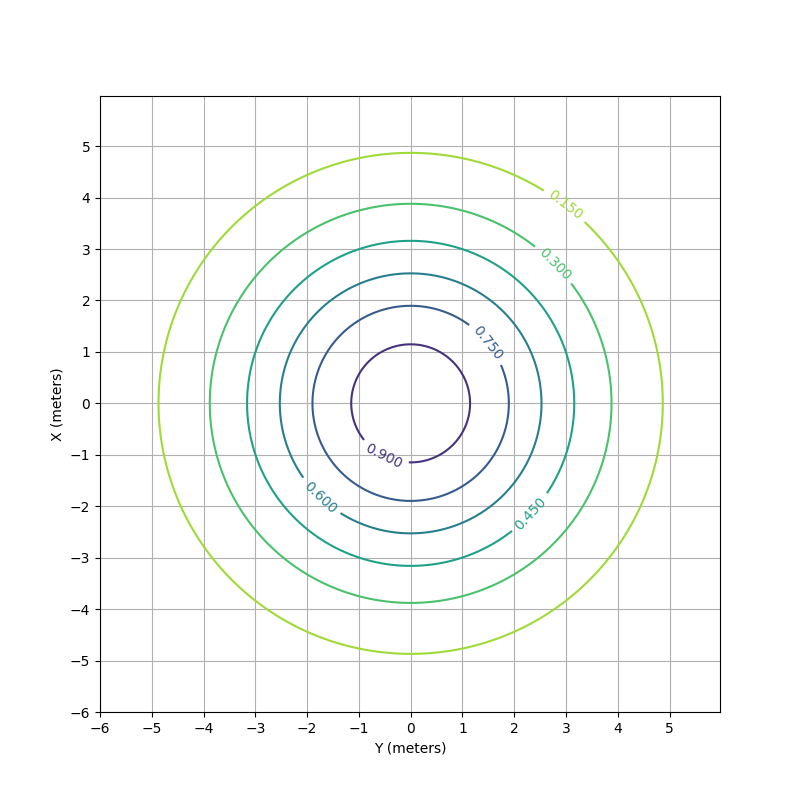}} 
\begin{figure}[]
\centering
    \begin{subfigure}{.49\linewidth}
        \centering
        \cookingFigure
        \caption{Cooking. \break \centering \(\sigma_{h}\) = 0.5, \(\sigma_{s}\) = 1.5, \(\sigma_{r}\) = 0.5}
        \label{fig:cooking_shape}
    \end{subfigure}
    \begin{subfigure}{.49\linewidth}
        \centering
        \runningFigure
        \caption{Running. \break \centering \(\sigma_{h}\) = 2.0, \(\sigma_{s}\) = 0.7, \(\sigma_{r}\) = 0.7}
        \label{fig:running_shape}
    \end{subfigure}
    \begin{subfigure}{.49\linewidth}
        \centering
        \standingFigure
        \caption{Standing. \break \centering \(\sigma_{h}\) = 0.8, \(\sigma_{s}\) = 1.0, \(\sigma_{r}\) = 1.0}
        \label{fig:standing_shape}
    \end{subfigure}
    \begin{subfigure}{.49\linewidth}
        \centering
        \bathroomFigure
        \caption{In the bathroom. \break \centering \(\sigma_{h}\) = 2.5, \(\sigma_{s}\) = 2.5, \(\sigma_{r}\) = 2.5}
        \label{fig:bathroom_shape}
    \end{subfigure}
\caption{Different proxemic shapes based on the context information.}
\label{fig:activity_prox}
\end{figure}

Using this model allows us to create areas around the people detected with different sizes and shapes. Figure \ref{fig:activity_prox} shows four activities' representations. If a person is cooking in the kitchen is expected, he/she is moving from right to left, going from the ceramic stove to the cut zone or the fridge. Using this representation, figure \ref{fig:cooking_shape}, a robot navigating in a domestic environment could pass behind the person, reducing the collision risk and improving his/her comfort. A similar situation is a person moving with a determined velocity in the robot's surroundings, figure \ref{fig:running_shape}. The velocity could be estimated, and the proxemic zones will be updated with this estimation, updating the \(\sigma_{h}\) parameter from the model. Thus, it creates a big zone in a person's front where navigate is forbidden or penalized, avoiding hit with a person in a hurry and with a dynamic size, based on the velocity estimation.
This tool also allows us to create different proxemics zones according to the human location. For example, a person in the bathroom could be no comfortable if a robot enters when he/she is in the shower or similar. Figure \ref{fig:bathroom_shape} shows how the proxemics zones have expanded to forbid the navigation in this area. Moreover, in this case, the orientation of the person is unknown. Because of this, the shape of this proxemics zones is like a standard Gaussian.


\subsection{Adaptation of the proxemic areas, the cooperation zone}
It can be argued that the use of proxemics zones is the most extent method to perform human awareness navigation, defining areas where the navigation is forbidden \cite{Kruse}. On the other hand, the robots have to behave socially and naturally and perform daily tasks with humans \cite{breazeal2004designing}. We can see how these two concepts collide, one restricts navigation around people, and the other promotes human-robot collaboration and interaction. It is necessary to create a mapping or a representation that takes into account the comfort and safety of people and at the same time allows the execution of tasks such as the robot approaching a person to give him a message or offer a information or the robot following or accompanying a person to a specific position.

\newcommand{\sLayerA}{\includegraphics[width=2.0in]{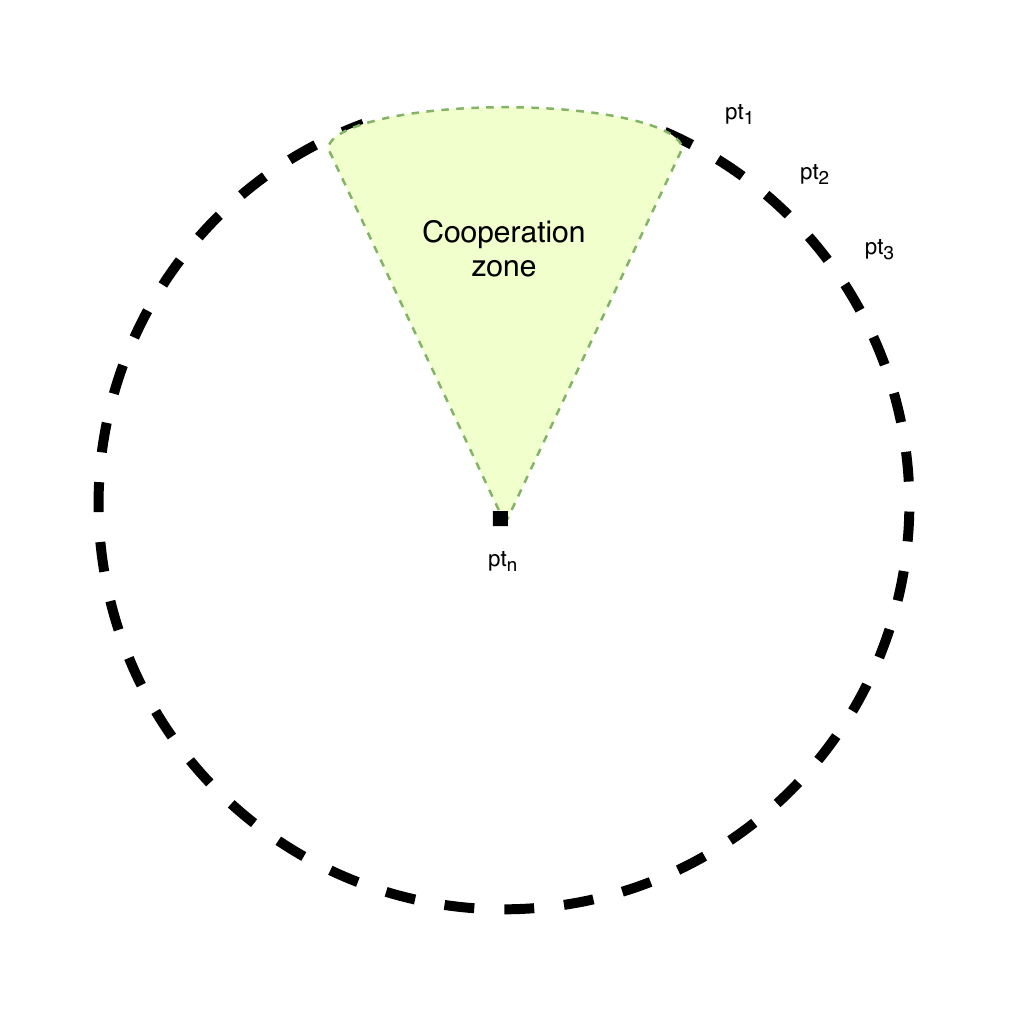}}
\newcommand{\sLayerB}{\includegraphics[width=2.0in]{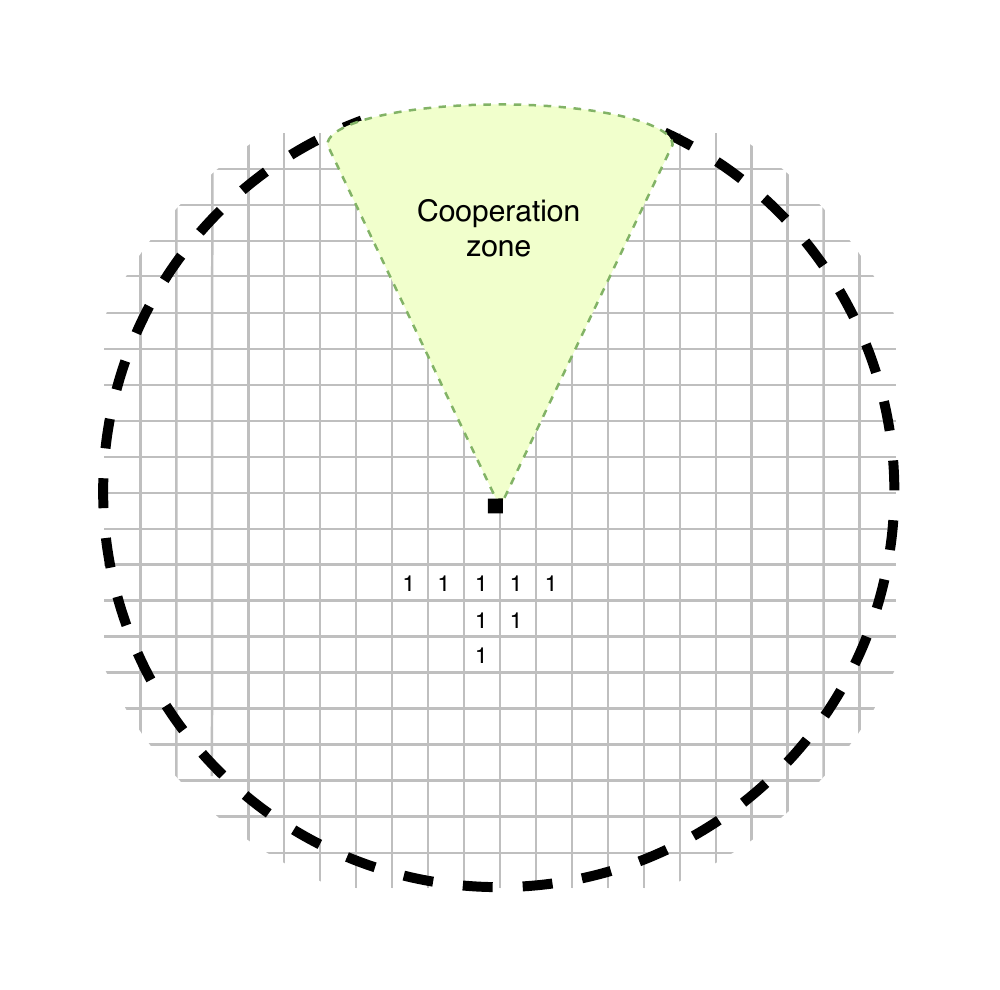}}
\newcommand{\sLayerC}{\includegraphics[width=2.0in]{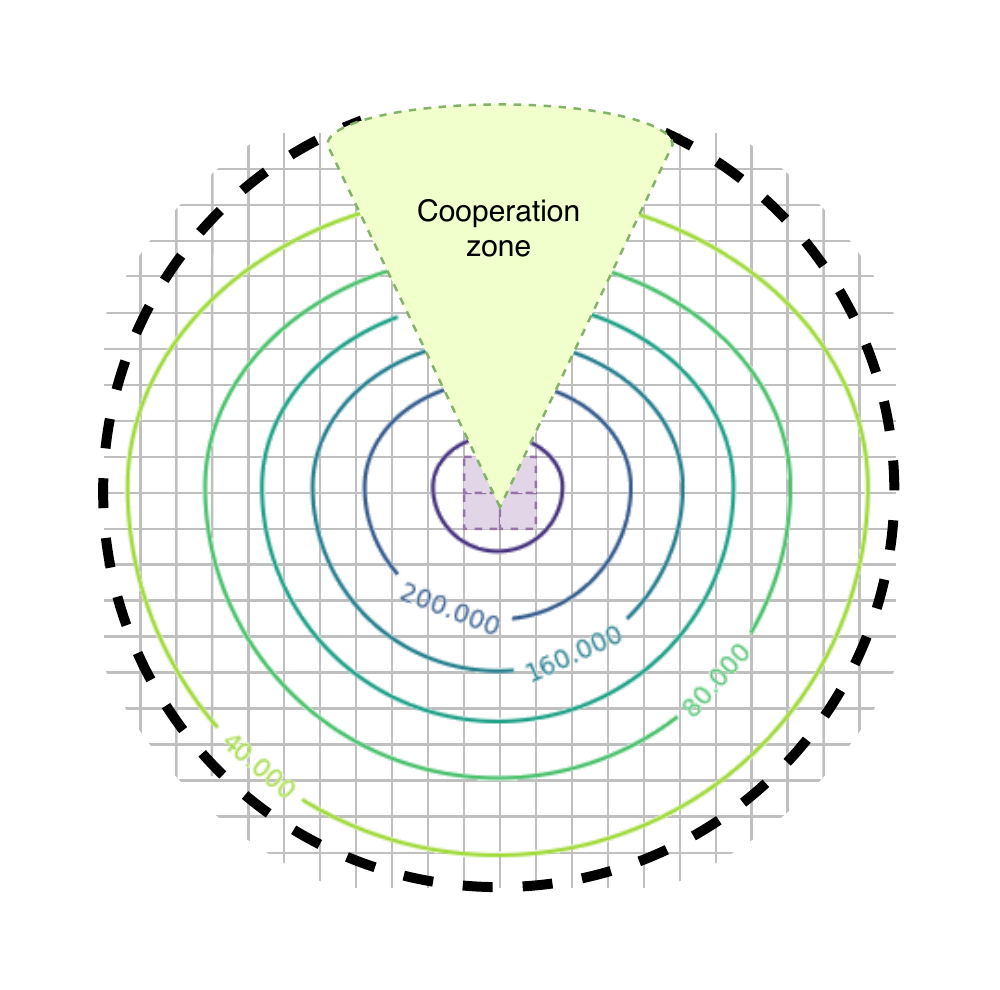}}
\newcommand{\sLayerD}{\includegraphics[width=2.0in]{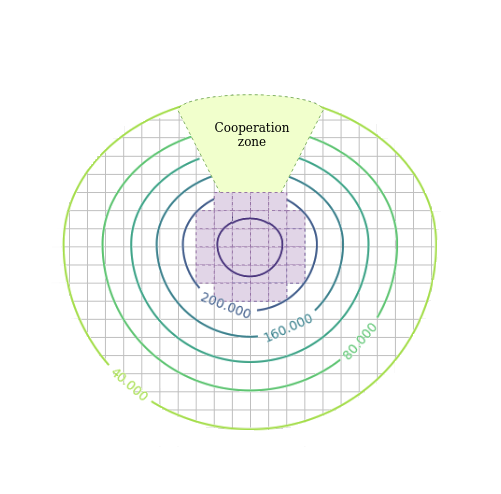}} 
\begin{figure}[]
\centering
    \begin{subfigure}{.49\linewidth}
        \centering
        \sLayerA
        \caption{The proxemic base zone is \break created  as a point's vector, \break excluding the cooperation zone.}
        \label{fig:sLayerA}
    \end{subfigure}
    \begin{subfigure}{.49\linewidth}
        \centering
        \sLayerB
        \caption{The map cells that are enclosed by the points defined in the previous step are searched.}
        \label{fig:sLayerB}
    \end{subfigure}
    \begin{subfigure}{.49\linewidth}
        \centering
        \sLayerC
        \caption{Each cell is evaluated with the Asymmetric Gaussian function defined.}
        \label{fig:sLayerC}
    \end{subfigure}
    \begin{subfigure}{.49\linewidth}
        \centering
        \sLayerD
        \caption{The intimate zone is added to the output of the previous step (purple cells). The intimate zone cells will be considered as an obstacle.}
        \label{fig:sLayerD}
    \end{subfigure}
\caption{Social layer workflow. From the human position to create its proxemics with a cooperation zone.}
\label{fig:cooperation_zone_prox}
\end{figure}

This article propose the creation of the cooperation zones, figure \ref{fig:cooperation_zone_prox}. As we see in the figure \ref{fig:sLayerD}, this cooperation zone is located outside the intimate zone and inside the personal zone, so the robot will keep a prudential distance to avoid colliding or discomforting the person and allowing a comfortable and natural interaction. This cooperation zone will be coded as a free zone on the map, so the navigation in it will be fluid.
The cooperation zones are configurable and can be set from 0 to 2 zones for each person. They can be located in the desired orientation and size, depending on the task for which they are designed. In this way, a cooperation zone to facilitate HRI tasks will be located in an orientation equal to the person or a cooperation zone designed to accompany people will be composed of two subzones, one on each side of the person.

\section{Building the social layer}
\label{sec:Social_layer}

One of the essential capabilities of a social robot is to move through space. It is necessary to integrate the sensory information obtained by the robot into a map to navigate through space safely and effectively. The previous reason, coupled with the widespread adoption by industry and the scientific community of ROS/ROS2 as a software development framework for robots, makes the ROS navigation \cite{5509725} one of the most widely used, robust, and stable packages on the platform. 
This navigation stack represents the sensory information on two grid maps or cost maps, global and local. The framework chosen for this research is ROS2, the most advanced version of ROS. In this one, the global cost map is used by the planner to calculate the path from robot current position to target, and the local cost map is used by the controller generating movements to follow this path, avoiding unexpected obstacles. Both the global cost map and the local cost map result from the combination of the different layers that compose it \cite{Lu2014}. Figure \ref{fig:nav_stack} shows how the default cost maps are split into different layers and which layers have been developed during this investigation, figure \ref{fig:proposedFig}. We will comment briefly on each layer's function to better understand how they work and what information they add to the final map.

\newcommand{\luFig}{\includegraphics[width=2.0in]{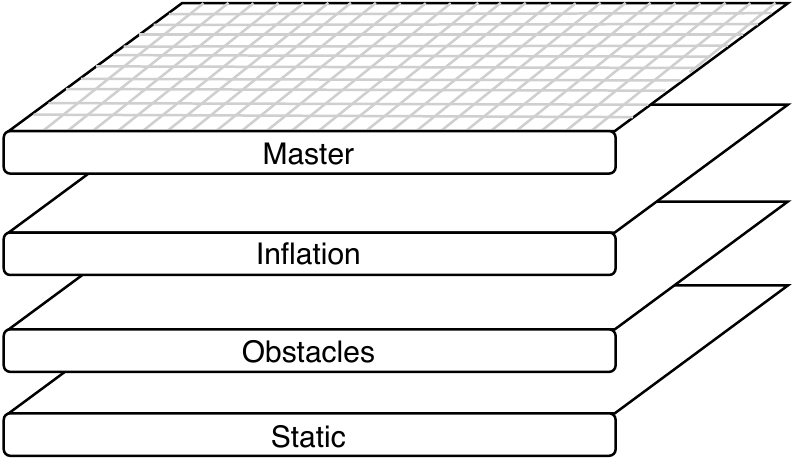}}
\newcommand{\proposedFig}{\includegraphics[width=2.0in]{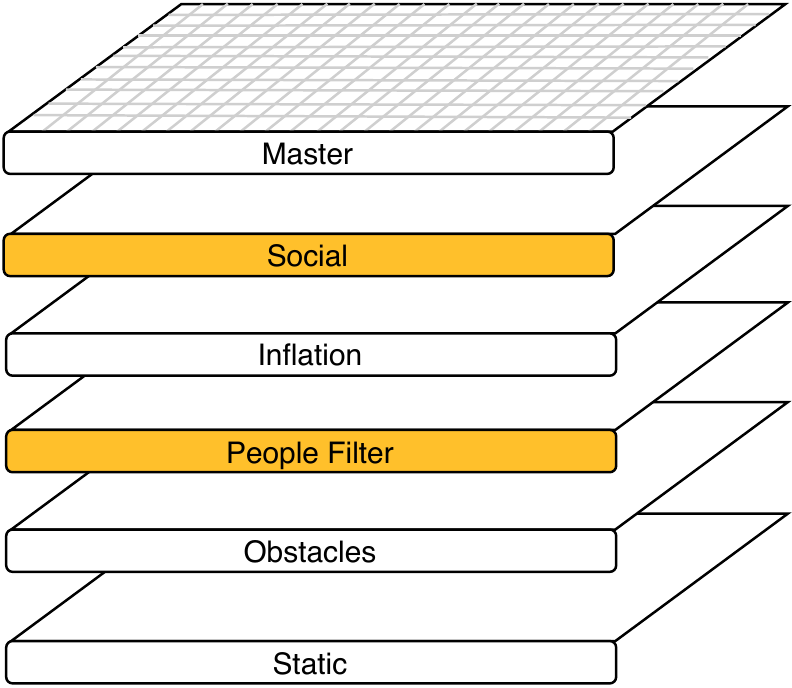}}
\begin{figure}[]
\centering
    \begin{subfigure}{.49\linewidth}
        \centering
        \luFig
        \caption{Default navigation stack.}
        \label{fig:luFig}
    \end{subfigure}
    \begin{subfigure}{.49\linewidth}
        \centering
        \proposedFig
        \caption{Social navigation stack.}
        \label{fig:proposedFig}
    \end{subfigure}
\caption{Previous and propose layers and their order.}
\label{fig:nav_stack}
\end{figure}

\begin{itemize}
    \item \textbf{Static:} The static layer adds the a priori information of the scenario. 
    \item \textbf{Obstacles:} The obstacle layer is subscribe to the sensors data (laser, ultrasounds, depth camera...) and materialise this information in the cost map. In addition to adding obstacle information, it also removes previously added information if the object disappears.
    \item \textbf{Inflation:} The inflation layer enlarges each of the obstacles previously added to the map by a radius equal to the robot's radius to ensure the safety of the robot's navigation. Also, the inflation process is made using a Gaussian function, so that the robot's speed is reduced when approaching an obstacle.
    \item \textbf{Master:} It is the final map and will be used by the planner and controller. It contains information about each one of the layers.
\end{itemize}

We can see how the order in which the layers are placed matters. It must be taken into account when developing a new layer to does not negatively affect the system's overall performance. For example, if we were to place the inflation layer before the obstacle layer, we would only be performing the process of expanding up obstacles to those represented on the scenario's map. The obstacles perceived by the sensors would be seen as points on the map, and it would be complicated to avoid them during navigation.

The layers developed in this research will be briefly presented below.

\begin{itemize}
    \item \textbf{People filter:} It is located after the obstacle layer and has as input the people's position on the map. At this point, the obstacle layer will have included the information of the people on the map, as they were obstacles. It is necessary to remove this information to have total control of the proxemic areas, so the cells around the people in a radius equal to the radius of their intimate area are marked as free cells.
    \item \textbf{Social:} It is located after the inflation layer and also has as input people's position. This layer adds the information of the people proxemic zones to the map. Section \ref{sec:Framework} shows in depth how these proxemic zones are.
\end{itemize}

As mentioned above, the developed layers need people's positions. This information can be obtained from a deep learning system, from a motion capture system, or, as in this article, from a simulator. It does not matter what system is used. The most important thing is that the information is represented correctly with the robot and the map. To do this, we will use the transform tree, tf2 \cite{6556373}, from ROS2. Robot's axes, sensors, map, and how each one is linked to other is represented in the tree. We will include in it the people's positions in the map frame. It will serve us as input for the two proposed layers.

Another essential tool that should be highlighted is the parameters to configure the proxemic zones. Through the ROS2 parameter system, we can set up new proxemic zones or configure, even during the execution of the system, the existing ones. This tool offers us great flexibility, necessary for the easy adapt of the system.

\section{Experimental results}
\label{sec:Experiments}
The experiments carried out are aimed at demonstrating the improvement in people's comfort by using the proposed system as opposed to the default navigation system. For this purpose, the metrics already established by the scientific community have been used \cite{10.1007/978-3-319-68345-4_25} \cite{okal2016learning}, formally described in \cite{Vega2019}: \(d_{min}\), average minimum distance to a human during navigation; \(d_{t}\), distance traveled; \(\tau\), navigation time; and \(Psi\), personal space intrusions.

Experiments have also been carried out to show the system's behavior during the execution of two collaborative tasks, escorting and following. 
ROS2Planning System \cite{ros2planning} has been used to implement each of the proposed actions.
It is an IA planning framework based on PDDL \cite{fox2003pddl2}, which uses popf \cite{coles2010forward} as planner.Using this tool, we can easily decompose complex tasks into a sequence of more uncomplicated actions. 

Another fundamental tool for performing the experiments is the pedestrian simulator based on the social force model, PedSim \cite{pedsimRos}\cite{gloor2016pedsim}.
This simulator will provide people's position and orientation in each instant of time.

The tests were performed on a computer with an Intel Core i7-8550U 1.8 GHz processor with 16 Gb of DDR4 RAM and Ubuntu GNU/Linux 18.04 using Gazebo as simulator and ROS2 Eloquent as robot framework.

\newcommand{\gazebo}{\includegraphics[width=2.5in]{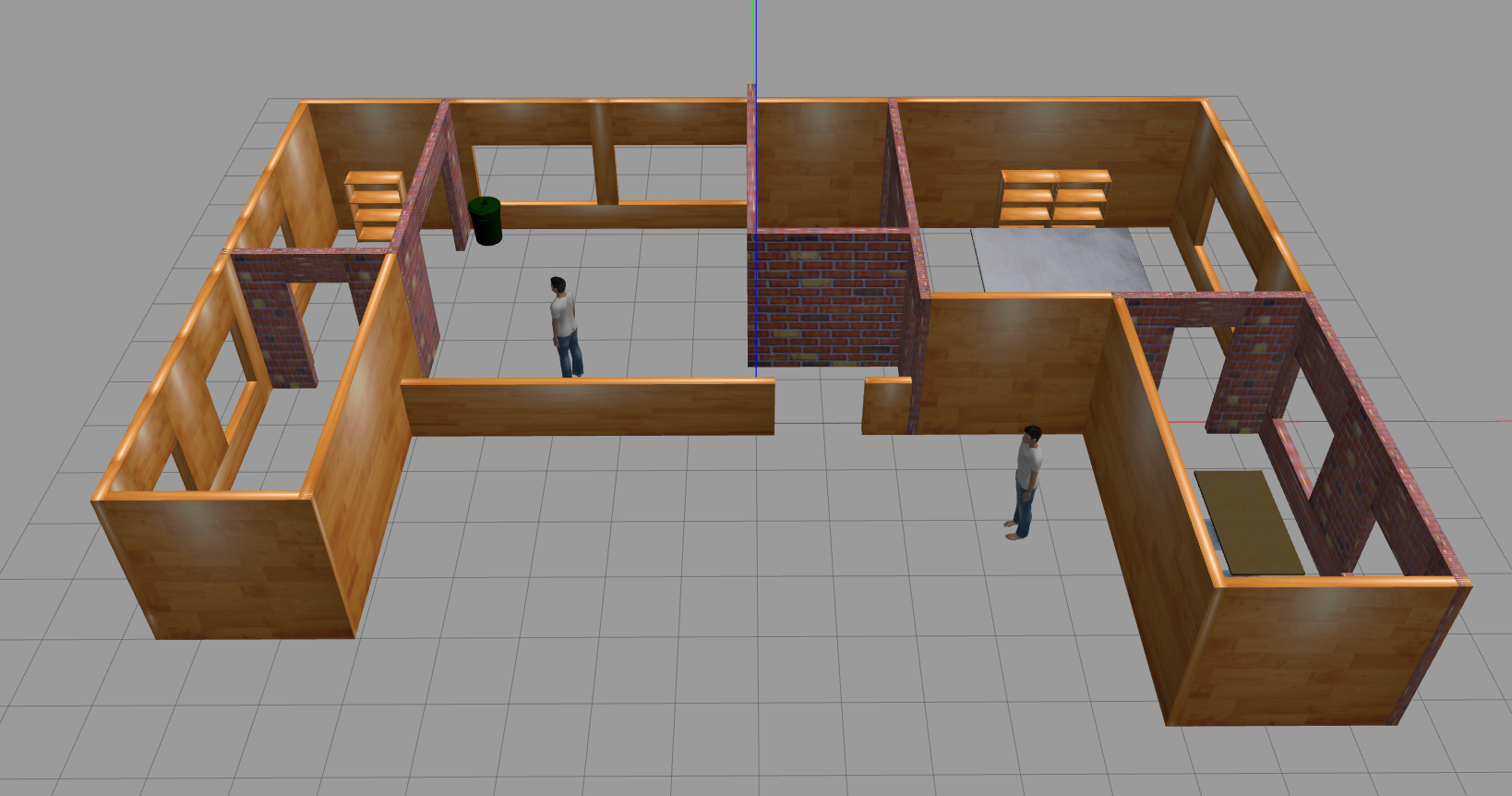}} 
\begin{figure}[]
    \centering
    \gazebo
    \caption{Domestic scenario from Gazebo simulator.}
    \label{fig:gazebo}
\end{figure}

\subsection{Approaching people}
\label{sec:experiments-Approaching}
In this experiment, we want to compare the behavior of the default navigation system of ROS2 besides the proposed system when performing an approaching task. The navigation speed was set at 0.3 m/s \cite{Butler2001}, the intimate zone radius at 0.4m, and the personal zone radius at 1.2m. Figure \ref{fig:approach_zones} shows the robot's representation of the person using the two systems. In the first case, we see how the robot does not differentiate people from other obstacles and therefore represents them on the map as such. On the other hand, we can see how the proxemic shape corresponding to the approach action has been established using the proposed system.

\newcommand{\approachZone}{\includegraphics[width=1.2in]{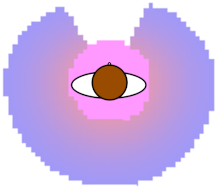}}
\newcommand{\defaultZone}{\includegraphics[width=1.2in]{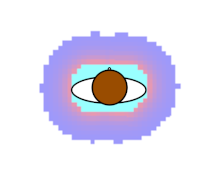}}
\begin{figure}[]
\centering
    \begin{subfigure}{.49\linewidth}
        \centering
        \defaultZone
        \caption{Person like an obstacle.}
        \label{fig:defaultZone}
    \end{subfigure}
    \begin{subfigure}{.49\linewidth}
        \centering
        \approachZone
        \caption{A proxemic zone with the cooperation area for HRI.}
        \label{fig:approachZone}
    \end{subfigure}
\caption{Representations of a person perceived by the robot.}
\label{fig:approach_zones}
\end{figure}

\newcommand{\approachActionPoints}{\includegraphics[width=1.5in]{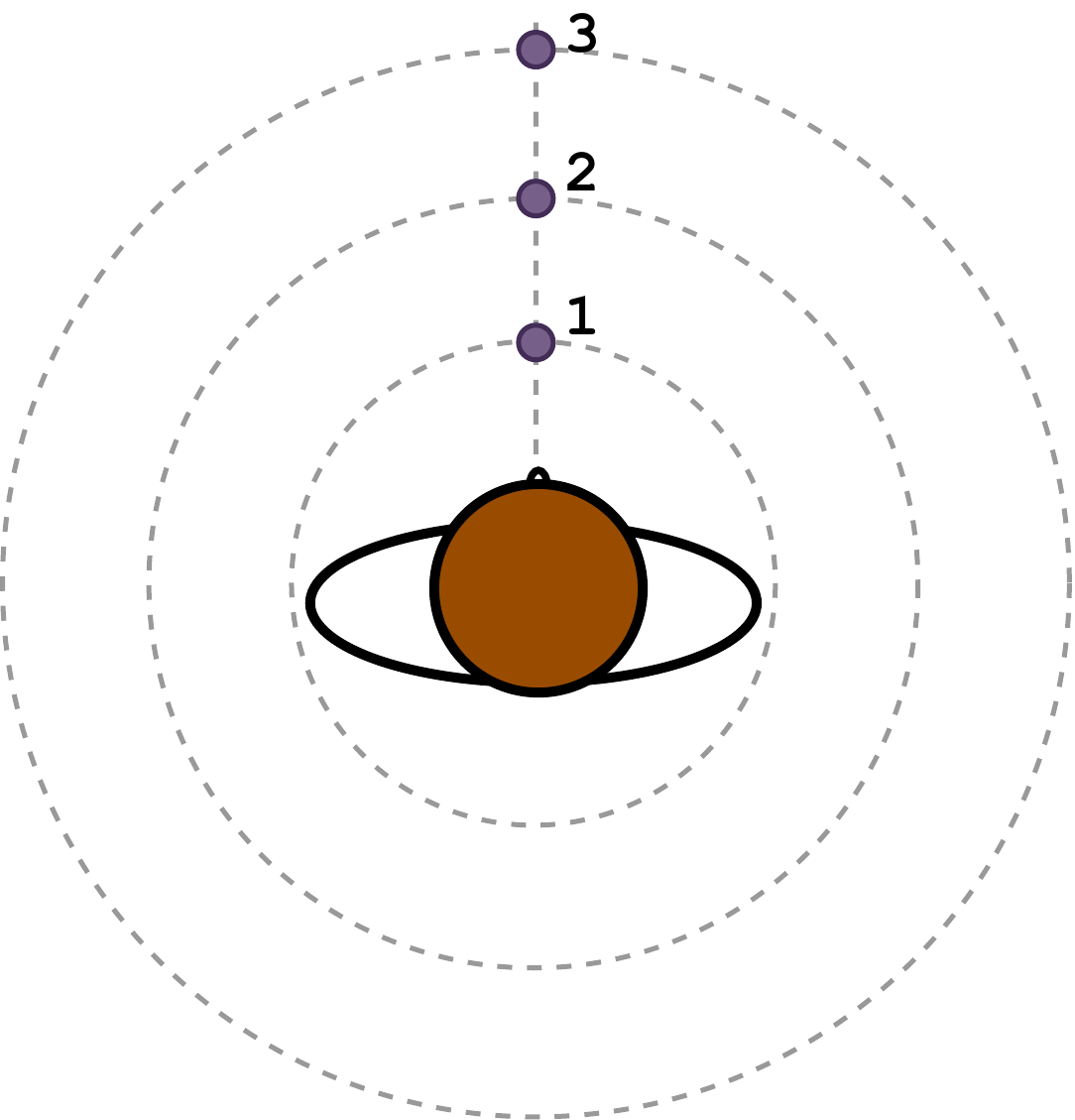}}
\newcommand{\escortActionPoints}{\includegraphics[width=1.5in]{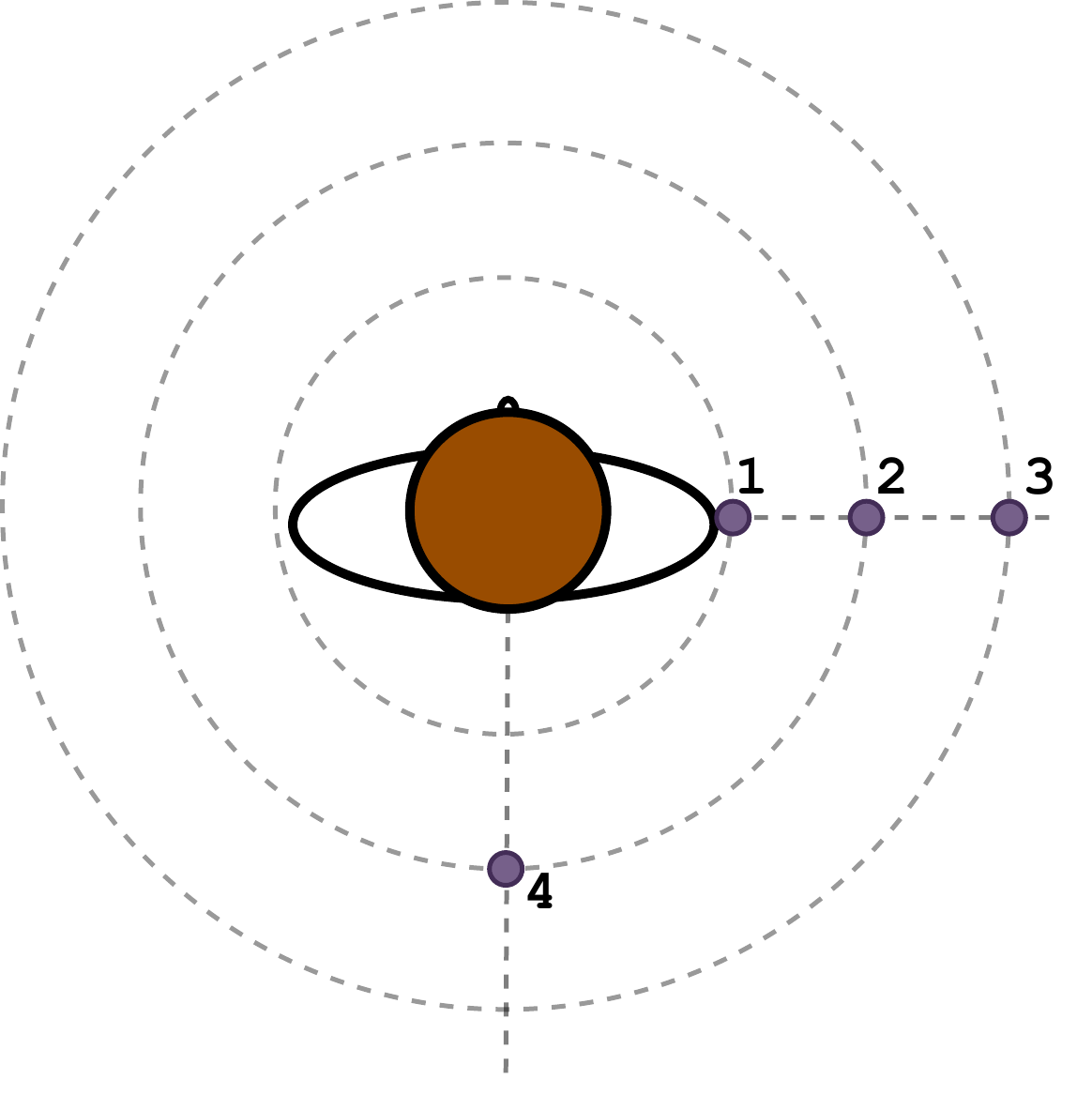}}
\begin{figure}[]
\centering
    \begin{subfigure}{.49\linewidth}
        \centering
        \approachActionPoints
        \caption{Approach action.}
        \label{fig:approachActionPoints}
    \end{subfigure}
    \begin{subfigure}{.49\linewidth}
        \centering
        \escortActionPoints
        \caption{Escort action.}
        \label{fig:escortActionPoints}
    \end{subfigure}
\caption{Navigation predefined points for the actions.}
\label{fig:action_points}
\end{figure}

The implementation of the approach action is described below. This action takes as input the position and orientation of the target person and, using an approach similar to that proposed by Koay et al. \cite{Koay2017}, a set of 3 predefined points are established as possible goals for the navigation system, figure \ref{fig:approachActionPoints}. The distance from these points to the human is as follows:
\begin{itemize}
    \item 1st: Intimate zone radius.
    \item 2nd: Intimate zone radius plus the robot's radius.
    \item 3rd: Intimate zone radius plus twice the robot's radius.
\end{itemize}
These points have an angle equal to the orientation of the person, although in the opposite direction. 
Once the points have been established, they are evaluated concerning the map in order of proximity to the person. Once the best point is obtained, it is sent to the navigation system.
With this implementation, the robot approaches the person from the front, regardless of the representation system we use.

Figure \ref{fig:gazebo} shows the scenario configuration for this experiment. We have the robot in the initial position and the person at a distance of 3.5m. Each iteration consists of 2 actions, \textit{approach} action, and \textit{return to home} action.  During the development of the experiment, the person's position is fixed, and its orientation changes in a random way in each iteration. Table \ref{table:1} shows the results obtained during the experiment, after the execution of 100 iterations.

\setlength{\tabcolsep}{12pt}
\captionsetup[table]{skip=3pt}
\begin{table}[h]
\centering
 \begin{tabular}{c c c} 
 \hline
 Parameter & ROS2 Navigation & Proposed approach\\ [0.5ex] 
 \hline\hline
   \(\tau (s)\) & 89.4 (22.02) & 89.28(17.64)  \\
 \hline
   \(d_{t} (m)\) & 7.06(1.37) & 7.71(1.67)\\ 
 \hline
   \(d_{min} (m)\) & 0.68(0.1) & 0.57(0.035)  \\
 \hline
   \(Psi(Personal) (\%)\) & 39.86(10.03) &  7.83(5.33) \\ [1ex] 
   \(Psi(Intimate) (\%)\) & 0(0) &  0(0) \\ [1ex]

\end{tabular}
\caption{Social navigation metrics for Approaching Test: for each parameter its mean and standard deviation are provided in parenthesis.}
\label{table:1}
\end{table}

We see how there has been a significant decrease in the personal area occupation percentage using the proposed system. Also, the minimum distance to people has been reduced, keeping in 0.54m - 0.60m shown in \cite{Walters2009} as adequate, from the human point of view, to carry out voice interaction tasks or human-robot handovers. The task execution time and distance traveled have been maintained, thus gaining comfort without affecting the system's overall performance.


\subsection{Collaborative actions}
\label{sec:experiments-Collaborative}

\newcommand{\defaultFootprint}{\includegraphics[width=1.0in]{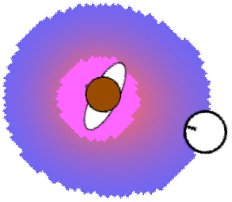}}
\newcommand{\defaultPath}{\includegraphics[width=3.3in]{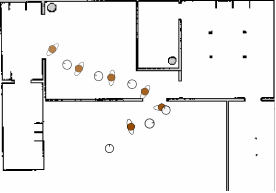}}
\newcommand{\escortFootprint}{\includegraphics[width=1.0in]{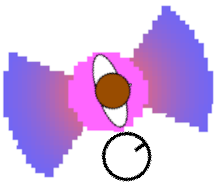}}
\newcommand{\escortPath}{\includegraphics[width=3.3in]{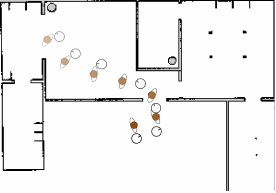}} 
\begin{figure}[]
\centering
    \begin{subfigure}{.29\linewidth}
        \centering
        \escortFootprint
        \caption{Escort proxemic shape.}
        \label{fig:escortFootprint}
    \end{subfigure}
    \begin{subfigure}{.69\linewidth}
        \centering
        \escortPath
        \caption{Escorting task path.}
        \label{fig:escortPath}
    \end{subfigure}
\caption{Robot is accompanying a human trough a door.}
\label{fig:escorting_task}
\end{figure}

This experiment shows the system's behavior during the execution of a collaborative task with a human, the escorting task.
The escorting action implementation is very similar to that shown in the previous experiment, figure \ref{fig:escortActionPoints}. Now target points are on the right of the person in addition to a point behind the person. This point is useful when, during the task execution, the human passes through a narrow area, such as a door. Figure \ref{fig:escortPath} shows the path taken by the person and the robot during the task.
Although the proxemic zone established during the whole execution of the task is the one shown in figure \ref{fig:escortFootprint}, it has been omitted in figure \ref{fig:escortPath} to help in the comprehension of the robot's behavior during the task.
If one looks in the narrow area of the path, the robot cannot continue next to the person, and it must go now behind them. When there is more space on the stage, the robot recovers its position and stands next to the person.

\newcommand{\escortDistance}{\includegraphics[width=3.0in]{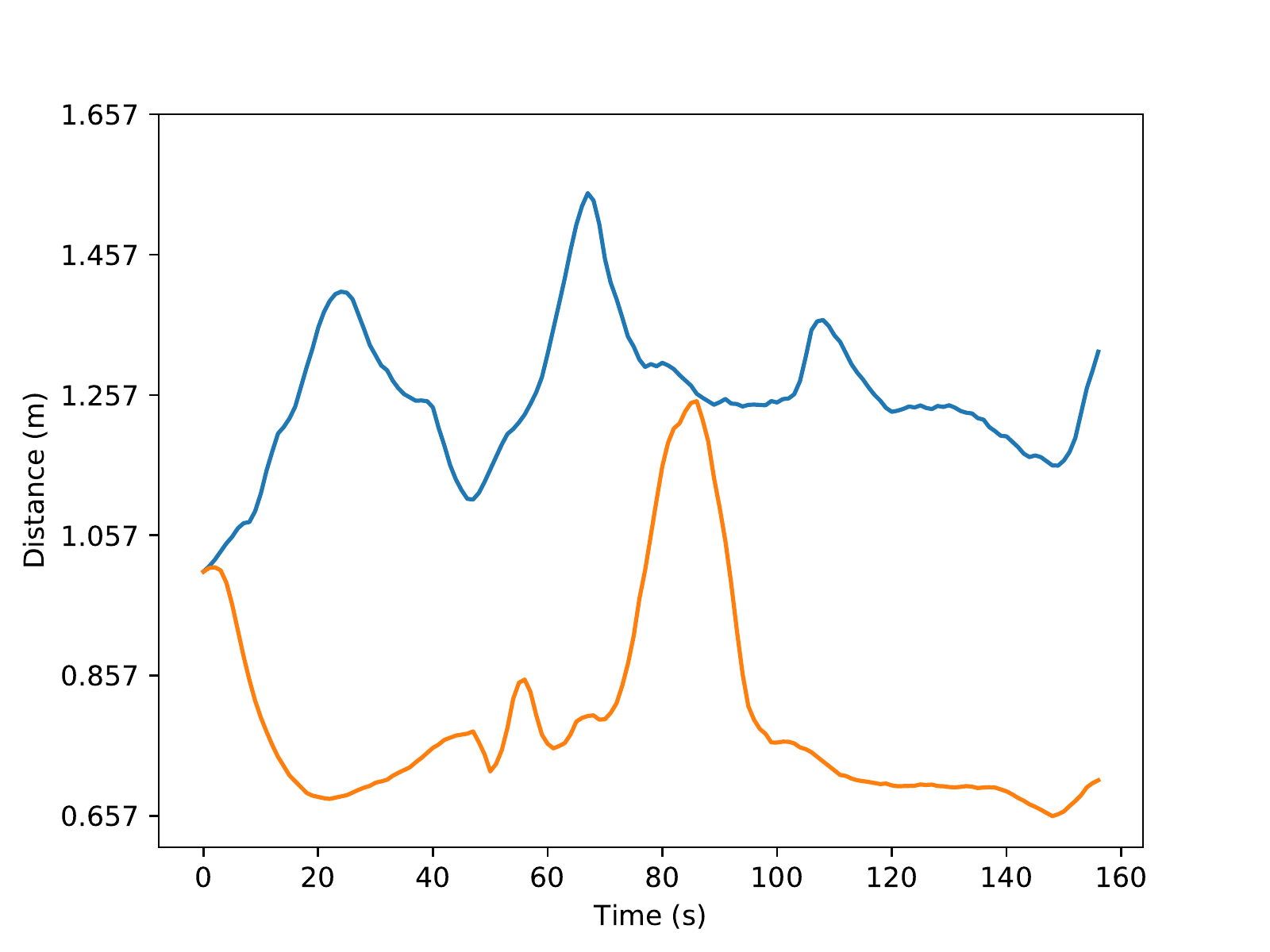}} 
\begin{figure}[]
    \centering
    \escortDistance
    \caption{Mean distance between robot and human during the escorting, (blue) using Lu et al \cite{6696579} approach; (orange) using our approach.}
    \label{fig:escorting_task}
\end{figure}

Finally, figure \ref{fig:escorting_task} shows a comparison of the distance between the robot and the human during the escorting task execution using a classical proxemic method, based on concentric circles Gaussian function, and the proposed method.
The distance between the person and the robot is always higher. This is due to the robot is always behind the person, since the target points are located in a restricted navigation area, the personal area.
%
%

%
\section{Conclusions and future works}
\label{sec:Conclusions}
Perception and context awareness systems attract more and more attention as they provide valuable information about the environment. Accordingly, a domestic robot can create a representation from people different than any other obstacle. It can be useful to perform collaborative navigation tasks between humans and robots or simply to facilitate human-robot interaction without affecting people's comfort. 

One of the ways to represent people is through the proxemics theory. It defines the space around a person as different zones with different sizes. These proxemic zones are usually static and regular. The paper at hand proposes the creation of proxemic zones adaptable to the context, human activity, or culture, in order to provide more comfort or safety, and also zones that facilitate the execution of collaborative tasks, with the creation of the so-called \textit{cooperation zone}. It allows performing tasks such as accompanying a person, following them, or approaching them simply and safely. To do this representation, we propose the development of open-source ROS2 navigation layers, \textit{people filter} and \textit{social layer}. 

The proposed framework can be tested and used by the scientific community since it is in a public repository, and it offers the necessary resources for the reproducibility of the results.

Future works include the experimentation and study of the solution in a real environment with real participants and the integration in a cognitive architecture that provides to the proposed system more information about the people or their behavior, arriving to learn of their habits or their daily activity automatically.

%
%
\bibliographystyle{styles/bibtex/spmpsci}      
\bibliography{bibliography}   

\end{document}